\documentclass[11pt,letterpaper]{article}
\usepackage{mypaper}
\usepackage{times}
\usepackage{latexsym}
\usepackage{enumerate}
\usepackage{graphicx}
\usepackage{multirow}

\naaclfinalcopy 
\def\naaclpaperid{703} 

\title{LSTM Neural Reordering Feature for Statistical Machine Translation}

\author{Yiming Cui, Shijin Wang\and Jianfeng Li\\
	    iFLYTEK Research, Beijing, China\\
	    {\tt \{ymcui,sjwang3,jfli3\}@iflytek.com}
	    }

\date{}

\begin{document}

\maketitle

\begin{abstract}
Artificial neural networks are powerful models, which have been widely applied into many aspects of machine translation, such as language modeling and translation modeling. Though notable improvements have been made in these areas, the reordering problem still remains a challenge in statistical machine translations. In this paper, we present a novel neural reordering model that directly models word pairs and their alignment. Further by utilizing LSTM recurrent neural networks, much longer context could be learned for reordering prediction. Experimental results on NIST OpenMT12 Arabic-English and Chinese-English 1000-best rescoring task show that our LSTM neural reordering feature is robust, and achieves significant improvements over various baseline systems.
\end{abstract}

\section{Introduction}
In statistical machine translation, the language model, translation model, and reordering model are the three most important components. Among these models, the reordering model plays an important role in phrase-based machine translation \cite{Koehn2004Statistical}, and it still remains a major challenge in current study.

In recent years, various phrase reordering methods have been proposed for phrase-based SMT systems, which can be classified into two broad categories:
\begin{enumerate}[(1)]
\item {\em Distance-based RM}: Penalize phrase displacements with respect to the degree of non-monotonicity \cite{Koehn2004Statistical}.
\item {\em Lexicalized RM}: Conditions reordering probabilities on current phrase pairs. According to the orientation determinants, lexicalized reordering model can further be classified into word-based RM \cite{tillman:2004:HLTNAACL}, phrase-based RM \cite{koehn-EtAl:2007:PosterDemo}, and hierarchical phrase-based RM \cite{galley-manning:2008:EMNLP}.
\end{enumerate}

Furthermore, some researchers proposed a reordering model that conditions both current and previous phrase pairs by utilizing recursive auto-encoders \cite{li-EtAl:2014:Coling3}.

In this paper, we propose a novel neural reordering feature by including longer context for predicting orientations. We utilize a long short-term memory recurrent neural network (LSTM-RNN) \cite{Graves1997Long}, and directly models word pairs to predict its most probable orientation. Experimental results on NIST OpenMT12 Arabic-English and Chinese-English translation show that our neural reordering model achieves significant improvements over various baselines in 1000-best rescoring task.

\section{Related Work}
Recently, various neural network models have been applied into machine translation. 

Feed-forward neural language model was first proposed by Bengio et al. \shortcite{Bengio2003A}, which was a breakthrough in language modeling. Mikolov et al. \shortcite{Mikolov2011Extensions} proposed to use recurrent neural network in language modeling, which can include much longer context history for predicting next word. Experimental results show that RNN-based language model significantly outperform standard feed-forward language model.

Devlin et al. \shortcite{devlin-EtAl:2014:P14-1} proposed a neural network joint model (NNJM) by conditioning both source and target language context for target word predicting. Though the network architecture is a simple feed-forward neural network, the results have shown significant improvements over state-of-the-art baselines.

Sundermeyer et al. \shortcite{sundermeyer-EtAl:2014:EMNLP2014} also put forward a neural translation model, by utilizing LSTM-based RNN and bidirectional RNN. By introducing bidirectional RNNs, the target word is conditioned on not only the history but also future source context, which forms a full source sentence for predicting target words.

Li et al. \shortcite{li-liu-sun:2013:EMNLP} proposed to use a recursive auto-encoder (RAE) to map each phrase pairs into continuous vectors, and handle reordering problems with a classifier. Also, they suggested that by both including current and previous phrase pairs to determine the phrase orientations could achieve further improvements in reordering accuracy \cite{li-EtAl:2014:Coling3}.

By far, we have noticed that this is the first time to use LSTM-RNN in reordering model. We could include much longer context information to determine phrase orientations using RNN architecture. Furthermore, by utilizing the LSTM units, the network is able to capture much longer range dependencies than standard RNNs. 

Because we need to record fixed length of history information in SMT decoding step, we only utilize our LSTM-RNN reordering model as a feature in 1000-best rescoring step. As word alignments are known after generating n-best list, it is possible to use LSTM-RNN reordering model to score each hypothesis.

\section{Lexicalized Reordering Model}
In traditional statistical machine translation, lexicalized reordering models have been widely used \cite{koehn-EtAl:2007:PosterDemo}. It considers alignments of current and previous phrase pairs to determine the orientation. 

Formally, when given source language sentence $f=\{f_{1}, ... , f_{n}\}$, target language sentence $e=\{e_{1}, ... , e_{n}\}$, and phrase alignment $a=\{a_{1}, ... , a_{n}\}$, the lexicalized reordering model can be illustrated in Equation 1, which only conditions on $a_{i-1}$ and $a_{i}$, i.e. previous and current alignment.
\begin{equation}p({\bf o}|{\bf e},{\bf f})=\prod_{i=1}^np(o_{i}|e_{i},f_{a_{i}},a_{i-1},a_{i}) \end{equation}

In Equation 1, the $o_{i}$ represents the set of phrase orientations. For example, in the most commonly used MSD-based orientation type, $o_{i}$ takes three values: M stands for {\em monotone}, S for {\em swap}, and D for {\em discontinuous}. The definition of MSD-based orientation is shown in Equation 2.
\begin{equation}
o_{i}=\left\{
\begin{array}{ccc}
M,& {a_{i}-a_{i-1}=1}\\
S,& {a_{i}-a_{i-1}=-1}\\
D,& {|a_{i}-a_{i-1}| \neq 1}
\end{array} \right.
\end{equation}

For other orientation types, such as LR and MSLR are also widely used, whose definition can be found on Moses official website \footnote{\tt http://www.statmt.org/moses/}.

Recent studies on reordering model suggest that by also conditioning previous phrase pairs can improve context sensitivity and reduce reordering ambiguity. 

\section{LSTM Neural Reordering Model}
In order to include more context information for determining reordering, we propose to use a recurrent neural network, which has been shown to perform considerably better than standard feed-forward architectures in sequence prediction \cite{Mikolov2011Extensions}. However, RNN with conventional back-propagation training suffers from gradient vanishing issues \cite{BengioSimardFrasconi94} .

Later, long short-term memory was proposed for solving gradient vanishing problem, and it could catch longer context than standard RNNs with sigmoid activation functions. In this paper, we adopt LSTM architecture for training neural reordering model.
\subsection{Training Data Processing}
For reducing model complexity and easy implementation, our neural reordering model is purely lexicalized and trained on word-level.

We will take LR orientation type for explanations, while other orientation types (MSD, MSLR) can be induced similarly. Given a sentence pair and its alignment information, we can induce the word-based reordering information by following steps. Note that, we always evaluate the model in the order of target sentence.
\begin{enumerate}[(1)]
\item If current target word is {\em one-to-one} alignment, then we can directly induce its orientations, i.e. $\langle left \rangle$ or $\langle right \rangle$.
\item If current source/target word is {\em one-to-many} alignment, then we judge its orientation by considering its first aligned target/source word, and the other aligned target/source words are annotated as $\langle follow \rangle$ reordering type, which means these word pairs inherent the orientation of previous word pair.
\item If current source/target word is not aligned to any target/source words, we introduce a $\langle null \rangle$ token in its opposite side, and annotate this word pair as $\langle follow \rangle$ reordering type.
\end{enumerate}

Figure 1 shows an example of data processing.
\begin{figure}[htbp]
  \centering
  \includegraphics[width=0.5\textwidth]{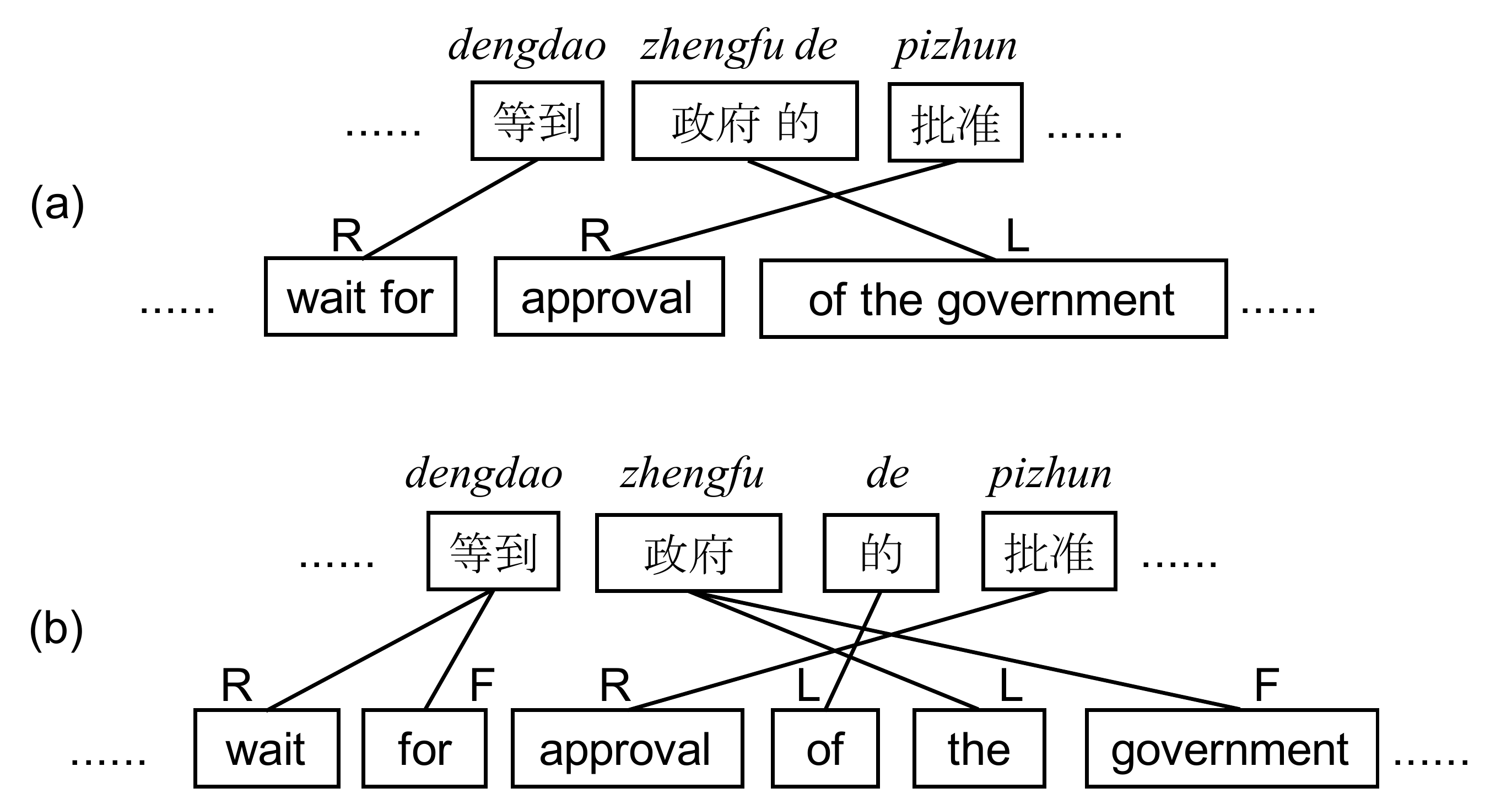}
  \caption{\label{font-table} Illustration of data processing. (a) Original reordering (alignment inside each phrase is omitted); (b) processed reordering, all alignments are regularized to word level. R-right, L-left, F-follow.}
\end{figure}

\subsection{LSTM Network Architecture}
After processing the training data, we can directly utilize the word pairs and its orientation to train a neural reordering model.

Given a word pair and its orientation, a neural reordering model can be illustrated by Equation 3.
\begin{equation}p({\bf o}|{\bf e},{\bf f})=\prod_{i=1}^np(o_{i}|e_{1}^{i},f_{1}^{a_{i}},a_{i-1},a_{i}) \end{equation}

Where $e_{1}^{i}=\{e_{1}, ... , e_{i}\}$, $f_{1}^{a_{i}}=\{f_{1}, ... , f_{a_{i}}\}$. Inclusion of history word pairs is done with recurrent neural network, which is known for its capability of learning history information. 

The architecture of LSTM-RNN reordering model is depicted in Figure 2, and corresponding equations are shown in Equation 4 to 6. 

\begin{equation}y_{i}=W_{1}*f_{a_{i}}+W_{2}*e_{i}\end{equation}
\begin{equation}z_{i}=LSTM(y_{i},W_{3},y_{1}^{i-1})\end{equation}
\begin{equation}p(o_{i}|e_{1}^{i},f_{1}^{a_{i}},a_{i-1},a_{i})=softmax(W_{4}*z_{i})\end{equation}

The input layer consists both source and target language word, which is in one-hot representation. Then we perform a linear transformation of input layer to a projection layer, which is also called embedding layer. We adopt extended-LSTM as our hidden layer implementation, which consists of three gating units, i.e. input, forget and output gates. We omit rather extensive LSTM equations here, which can be found in \cite{Graves2005Framewise}. The output layer is composed by orientation types. For example, in LR condition, the output layer contains two units: $\langle left \rangle$ and $\langle right \rangle$ orientation. Finally, we apply softmax function to obtain normalized probabilities of each orientation.

\begin{figure}[htbp]
  \centering
  \includegraphics[width=0.45\textwidth]{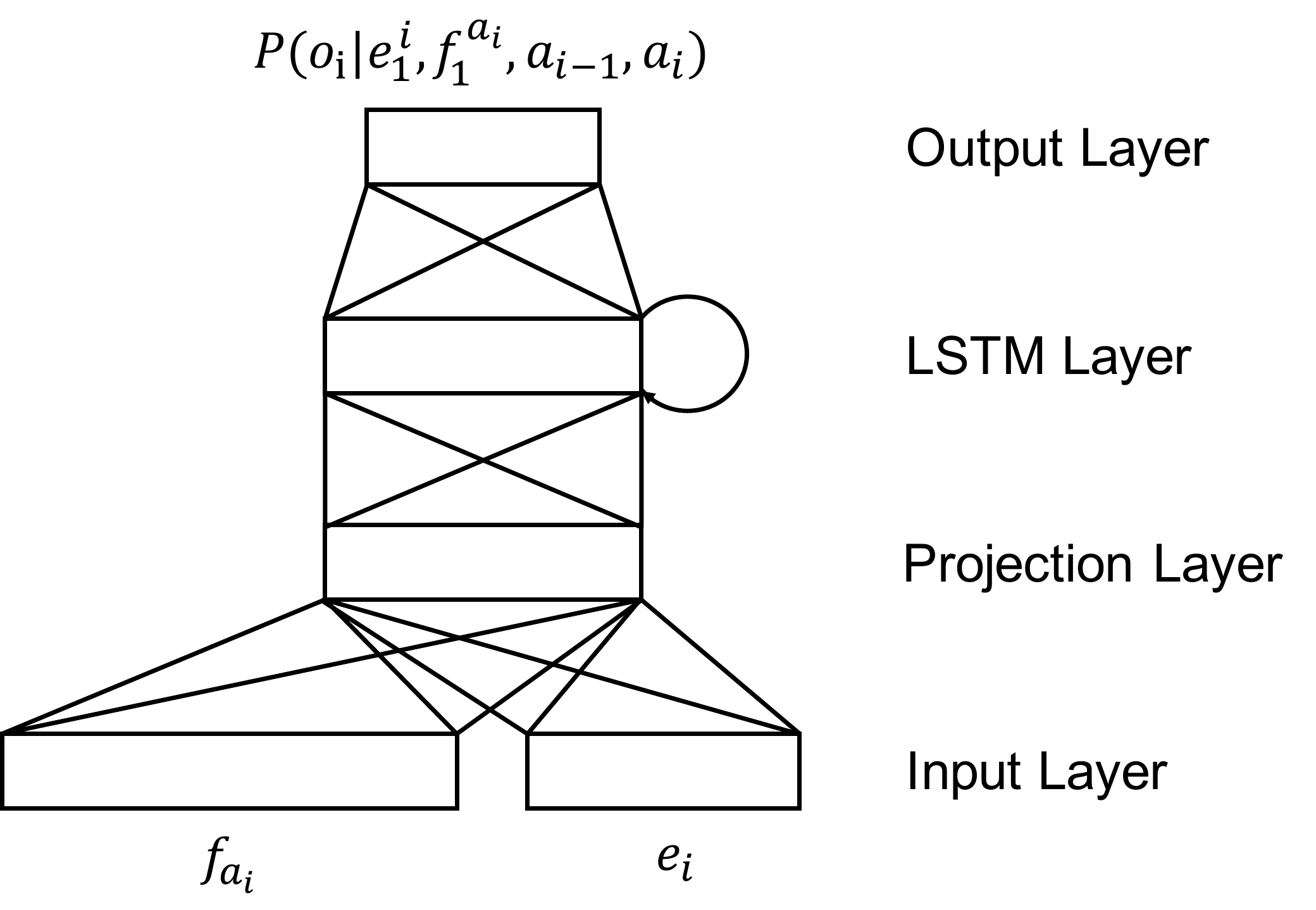}
  \caption{\label{font-table} Architecture of LSTM neural reordering model.}
\end{figure}

\section{Experiments}
\subsection{Setups}
We mainly tested our approach on Arabic-English and Chinese-English translation. The training corpus contains 7M words for Arabic, and 4M words for Chinese, which is selected from NIST OpenMT12 parallel dataset. We use the SAMA tokenizer\footnote{\tt https://catalog.ldc.upenn.edu/LDC2010L01} for Arabic word tokenization, and in-house segmenter for Chinese words. The English part of parallel data is tokenized and lowercased. All development and test sets have 4 references for each segment. The statistics of development and test sets are shown in Table 1. 

\begin{table}[tbp]
\centering
\begin{tabular}{|c|c|c|c|}
\hline \bf System & {\bf Dev} & {\bf Test1} & {\bf Test2} \\ 
\hline
\multirow{2}{1cm}{Ar-En} & MT04-05-06 & MT08 & MT09 \\
 & (3795) & (1360) & (1313) \\
\hline
\multirow{2}{1cm}{Zh-En} & MT05-08 & MT08.prog & MT12.rd \\
 & (2439) & (1370) & (820) \\
\hline
\end{tabular}
\caption{\label{font-table} Statistics of development and test set. The number of segments are indicated in brackets.}
\end{table}

The baseline systems are built with the open-source phrase-based SMT toolkit Moses \cite{koehn-EtAl:2007:PosterDemo}. Word alignment and phrase extraction are done by GIZA++ \cite{Och2000A} with L0-normalization \cite{vaswani-huang-chiang:2012:ACL2012}, and grow-diag-final refinement rule \cite{Koehn2004Statistical}. Monolingual part of training data is used to train a 5-gram language model using SRILM \cite{Stolcke2002Srilm}. Parameter tuning is done by K-best MIRA \cite{cherry-foster:2012:NAACL-HLT}. For guarantee of result stability, we tune every system 5 times independently, and take the average BLEU score \cite{clark-EtAl:2011:ACL-HLT2011}. The translation quality is evaluated by case-insensitive BLEU-4 metric \cite{papineni-EtAl:2002:ACL}. The statistical significance test is also carried out with paired bootstrap resampling method with $p<0.001$ intervals \cite{koehn:2004:EMNLP}. Our models are evaluated in a 1000-best rescoring step, and all features in 1000-best list as well as LSTM-RNN reordering feature are retuned via K-best MIRA algorithm.

For neural network training, we use all parallel text in the baseline training. As a trade-off between computational cost and performance, the projection layer and hidden layer are set to 100, which is enough for our task (We have not seen significant gains when increasing dimensions greater than 100). We use an initial learning rate of 0.01 with standard SGD optimization without momentum. We trained model for a total of 10 epochs with cross-entropy criterion. Input and output vocabulary are set to 100K and 50K respectively, and all out-of-vocabulary words are mapped to a $\langle unk \rangle$ token.

\subsection{Results on Different Orientation Types}
At first, we test our neural reordering model (NRM) on the baseline that contains word-based reordering model with LR orientation. The results are shown in Table 2 and 3. 

As we can see that, among various orientation types (LR, MSD, MSLR), our model could give consistent improvements over baseline system. The overall BLEU improvements range from 0.42 to 0.79 for Arabic-English, and 0.31 to 0.72 for Chinese-English systems. All neural results are significantly better than baselines ($p<0.001$ level).

In the meantime, we also find that ``Left-Right'' based orientation methods, such as LR and MSLR, consistently outperform MSD-based orientations. The may caused by non-separability problem, which means that MSD-based methods are vulnerable to the change of context, and weak in resolving reordering ambiguities. Similar conclusion can be found in Li et al. \shortcite{li-EtAl:2014:Coling3} .

\begin{table}[htbp]
\centering
\begin{tabular}{|l|c|c|c|}
\hline {\bf Ar-En System} & {\bf Dev} & {\bf Test1} & {\bf Test2} \\ 
\hline
Baseline & 43.87 & 39.84 & 42.05 \\
+NRM\_LR & 44.43 & 40.53 & \bf 42.84 \\
+NRM\_MSD & 44.29 & 40.41 & 42.62 \\
+NRM\_MSLR & \bf 44.52 & \bf 40.59 & 42.78 \\
\hline
\end{tabular}
\caption{\label{font-table} LSTM reordering model with different orientation types for Arabic-English system.}
\end{table}

\begin{table}[htbp]
\centering
\begin{tabular}{|l|c|c|c|}
\hline {\bf Zh-En System} & {\bf Dev} & {\bf Test1} & {\bf Test2} \\ 
\hline
Baseline & 27.18 & 26.17 & 24.04 \\
+NRM\_LR & \bf 27.90 & 26.58 & \bf 24.70 \\
+NRM\_MSD & 27.49 & 26.51 & 24.39 \\
+NRM\_MSLR & 27.82 & \bf 26.78 & 24.53 \\
\hline
\end{tabular}
\caption{\label{font-table} LSTM reordering model with different orientation types for Chinese-English system.}
\end{table}

\subsection{Results on Different Reordering Baselines}
We also test our approach on various baselines, which either contains word-based, phrase-based, or hierarchical phrase-based reordering model. We only show the results of MSLR orientation, which is relatively superior than others according to the results in Section 5.2.

In Table 4 and 5, we can see that though we add a strong hierarchical phrase-based reordering model in the baseline, our model can still bring a maximum gain of 0.59 BLEU score, which suggest that our model is applicable and robust in various circumstances. However, we have noticed that the gains in Arabic-English system is relatively greater than that in Chinese-English system. This is probably because hierarchical reordering features tend to work better for Chinese words, and thus our model will bring little remedy to its baseline.

\begin{table}
\centering
\begin{tabular}{|l|c|c|c|}
\hline {\bf Ar-En System} & {\bf Dev} & {\bf Test1} & {\bf Test2} \\ 
\hline
Baseline\_wbe & 43.87 & 39.84 & 42.05 \\
+NRM\_MSLR & 44.52 & 40.59 & 42.78 \\
\hline   
Baseline\_phr & 44.11 & 40.09 & 42.21 \\
+NRM\_MSLR & 44.52 & 40.73 & 42.89 \\
\hline
Baseline\_hier & 44.30 & 40.23 & 42.38 \\
+NRM\_MSLR & 44.61 & 40.82 & 42.86 \\
\hline
\end{tabular}

\begin{tabular}{|l|c|c|c|}
\hline {\bf Zh-En System} & {\bf Dev} & {\bf Test1} & {\bf Test2} \\ 
\hline
Baseline\_wbe & 27.18 & 26.17 & 24.04 \\
+NRM\_MSLR & 27.90 & 26.58 & 24.70 \\
\hline   
Baseline\_phr & 27.33 & 26.05 & 24.13 \\
+NRM\_MSLR & 27.86 & 26.46 & 24.73 \\
\hline
Baseline\_hier & 27.56 & 26.29 & 24.38 \\
+NRM\_MSLR & 28.02 & 26.49 & 24.67 \\
\hline
\end{tabular}

\caption{\label{font-table} Results on various baselines for Arabic-English and Chinese-English system. ``wbe'': word-based; ``phr'': phrase-based; ``hier'': hierarchical phrase-based reordering model. All NRM results are significantly better than baselines ($p<0.001$ level).}
\end{table}

\section{Conclusions}
We present a novel work that build a reordering model using LSTM-RNN, which is much sensitive to the change of context and introduce rich context information for reordering prediction. Furthermore, the proposed model is purely lexicalized and straightforward, which is easy to realize. Experimental results on 1000-best rescoring show that our neural reordering feature is robust, and could give consistent improvements over various baseline systems.

In future, we are planning to extend our word-based LSTM reordering model to phrase-based reordering model, in order to dissolve much more ambiguities and improve reordering accuracy. Furthermore, we are also going to integrate our neural reordering model into neural machine translation systems.

\section*{Acknowledgments}

We sincerely thank the anonymous reviewers for their thoughtful comments on our work.

\bibliography{mypaper}

\begin{thebibliography}{}

\bibitem[\protect\citename{Bengio \bgroup et al.\egroup
  }1994]{BengioSimardFrasconi94}
Y.~Bengio, P.~Simard, and P.~Frasconi.
\newblock 1994.
\newblock Learning long-term dependencies with gradient descent is difficult.
\newblock {\em IEEE Transactions on Neural Networks}, 5(2):157--166.

\bibitem[\protect\citename{Bengio \bgroup et al.\egroup }2003]{Bengio2003A}
Yoshua Bengio, Holger Schwenk, Jean~Sébastien Senécal, Fréderic Morin, and
  Jean~Luc Gauvain.
\newblock 2003.
\newblock A neural probabilistic language model.
\newblock {\em Journal of Machine Learning Research}, 3(6):1137--1155.

\bibitem[\protect\citename{Cherry and
  Foster}2012]{cherry-foster:2012:NAACL-HLT}
Colin Cherry and George Foster.
\newblock 2012.
\newblock Batch tuning strategies for statistical machine translation.
\newblock In {\em Proceedings of the 2012 Conference of the North American
  Chapter of the Association for Computational Linguistics: Human Language
  Technologies}, pages 427--436, Montr\'{e}al, Canada, June. Association for
  Computational Linguistics.

\bibitem[\protect\citename{Clark \bgroup et al.\egroup
  }2011]{clark-EtAl:2011:ACL-HLT2011}
Jonathan~H. Clark, Chris Dyer, Alon Lavie, and Noah~A. Smith.
\newblock 2011.
\newblock Better hypothesis testing for statistical machine translation:
  Controlling for optimizer instability.
\newblock In {\em Proceedings of the 49th Annual Meeting of the Association for
  Computational Linguistics: Human Language Technologies}, pages 176--181,
  Portland, Oregon, USA, June. Association for Computational Linguistics.

\bibitem[\protect\citename{Devlin \bgroup et al.\egroup
  }2014]{devlin-EtAl:2014:P14-1}
Jacob Devlin, Rabih Zbib, Zhongqiang Huang, Thomas Lamar, Richard Schwartz, and
  John Makhoul.
\newblock 2014.
\newblock Fast and robust neural network joint models for statistical machine
  translation.
\newblock In {\em Proceedings of the 52nd Annual Meeting of the Association for
  Computational Linguistics (Volume 1: Long Papers)}, pages 1370--1380,
  Baltimore, Maryland, June. Association for Computational Linguistics.

\bibitem[\protect\citename{Galley and Manning}2008]{galley-manning:2008:EMNLP}
Michel Galley and Christopher~D. Manning.
\newblock 2008.
\newblock A simple and effective hierarchical phrase reordering model.
\newblock In {\em Proceedings of the 2008 Conference on Empirical Methods in
  Natural Language Processing}, pages 848--856, Honolulu, Hawaii, October.
  Association for Computational Linguistics.

\bibitem[\protect\citename{Graves and Schmidhuber}2005]{Graves2005Framewise}
A.~Graves and J.~Schmidhuber.
\newblock 2005.
\newblock Framewise phoneme classification with bidirectional lstm networks.
\newblock In {\em Proceedings in 2005 IEEE International Joint Conference on
  Neural Networks}, pages 2047--2052 vol. 4.

\bibitem[\protect\citename{Graves}1997]{Graves1997Long}
Alex Graves.
\newblock 1997.
\newblock Long short-term memory.
\newblock {\em Neural Computation}, 9(8):1735--1780.

\bibitem[\protect\citename{Koehn \bgroup et al.\egroup
  }2004]{Koehn2004Statistical}
Philipp Koehn, Franz~Josef Och, and Daniel Marcu.
\newblock 2004.
\newblock Statistical phrase-based translation.
\newblock In {\em Conference of the North American Chapter of the Association
  for Computational Linguistics on Human Language Technology-volume}, pages
  127--133.

\bibitem[\protect\citename{Koehn \bgroup et al.\egroup
  }2007]{koehn-EtAl:2007:PosterDemo}
Philipp Koehn, Hieu Hoang, Alexandra Birch, Chris Callison-Burch, Marcello
  Federico, Nicola Bertoldi, Brooke Cowan, Wade Shen, Christine Moran, Richard
  Zens, Chris Dyer, Ondrej Bojar, Alexandra Constantin, and Evan Herbst.
\newblock 2007.
\newblock Moses: Open source toolkit for statistical machine translation.
\newblock In {\em Proceedings of the 45th Annual Meeting of the Association for
  Computational Linguistics Companion Volume Proceedings of the Demo and Poster
  Sessions}, pages 177--180, Prague, Czech Republic, June. Association for
  Computational Linguistics.

\bibitem[\protect\citename{Koehn}2004]{koehn:2004:EMNLP}
Philipp Koehn.
\newblock 2004.
\newblock Statistical significance tests for machine translation evaluation.
\newblock In Dekang Lin and Dekai Wu, editors, {\em Proceedings of EMNLP 2004},
  pages 388--395, Barcelona, Spain, July. Association for Computational
  Linguistics.

\bibitem[\protect\citename{Li \bgroup et al.\egroup
  }2013]{li-liu-sun:2013:EMNLP}
Peng Li, Yang Liu, and Maosong Sun.
\newblock 2013.
\newblock Recursive autoencoders for {ITG}-based translation.
\newblock In {\em Proceedings of the 2013 Conference on Empirical Methods in
  Natural Language Processing}, pages 567--577, Seattle, Washington, USA,
  October. Association for Computational Linguistics.

\bibitem[\protect\citename{Li \bgroup et al.\egroup
  }2014]{li-EtAl:2014:Coling3}
Peng Li, Yang Liu, Maosong Sun, Tatsuya Izuha, and Dakun Zhang.
\newblock 2014.
\newblock A neural reordering model for phrase-based translation.
\newblock In {\em Proceedings of COLING 2014, the 25th International Conference
  on Computational Linguistics: Technical Papers}, pages 1897--1907, Dublin,
  Ireland, August. Dublin City University and Association for Computational
  Linguistics.

\bibitem[\protect\citename{Mikolov \bgroup et al.\egroup
  }2011]{Mikolov2011Extensions}
T.~Mikolov, S.~Kombrink, L.~Burget, and J.~H. Cernocky.
\newblock 2011.
\newblock Extensions of recurrent neural network language model.
\newblock In {\em IEEE International Conference on Acoustics, Speech and Signal
  Processing}, pages 5528--5531.

\bibitem[\protect\citename{Och and Ney}2000]{Och2000A}
Franz~Josef Och and Hermann Ney.
\newblock 2000.
\newblock A comparison of alignment models for statistical machine translation.
\newblock In {\em Proceedings of the 18th conference on Computational
  linguistics - Volume 2}, pages 1086--1090.

\bibitem[\protect\citename{Papineni \bgroup et al.\egroup
  }2002]{papineni-EtAl:2002:ACL}
Kishore Papineni, Salim Roukos, Todd Ward, and Wei-Jing Zhu.
\newblock 2002.
\newblock Bleu: a method for automatic evaluation of machine translation.
\newblock In {\em Proceedings of 40th Annual Meeting of the Association for
  Computational Linguistics}, pages 311--318, Philadelphia, Pennsylvania, USA,
  July. Association for Computational Linguistics.

\bibitem[\protect\citename{Stolcke}2002]{Stolcke2002Srilm}
Andreas Stolcke.
\newblock 2002.
\newblock Srilm --- an extensible language modeling toolkit.
\newblock In {\em Proceedings of the 7th International Conference on Spoken
  Language Processing (ICSLP 2002)}, pages 901--904.

\bibitem[\protect\citename{Sundermeyer \bgroup et al.\egroup
  }2014]{sundermeyer-EtAl:2014:EMNLP2014}
Martin Sundermeyer, Tamer Alkhouli, Joern Wuebker, and Hermann Ney.
\newblock 2014.
\newblock Translation modeling with bidirectional recurrent neural networks.
\newblock In {\em Proceedings of the 2014 Conference on Empirical Methods in
  Natural Language Processing (EMNLP)}, pages 14--25, Doha, Qatar, October.
  Association for Computational Linguistics.

\bibitem[\protect\citename{Tillman}2004]{tillman:2004:HLTNAACL}
Christoph Tillman.
\newblock 2004.
\newblock A unigram orientation model for statistical machine translation.
\newblock In Daniel~Marcu Susan~Dumais and Salim Roukos, editors, {\em
  HLT-NAACL 2004: Short Papers}, pages 101--104, Boston, Massachusetts, USA,
  May 2 - May 7. Association for Computational Linguistics.

\bibitem[\protect\citename{Vaswani \bgroup et al.\egroup
  }2012]{vaswani-huang-chiang:2012:ACL2012}
Ashish Vaswani, Liang Huang, and David Chiang.
\newblock 2012.
\newblock Smaller alignment models for better translations: Unsupervised word
  alignment with the l0-norm.
\newblock In {\em Proceedings of the 50th Annual Meeting of the Association for
  Computational Linguistics (Volume 1: Long Papers)}, pages 311--319, Jeju
  Island, Korea, July. Association for Computational Linguistics.

\end{thebibliography}
\bibliographystyle{mypaper}

\end{document}